\documentclass{article}

\usepackage{PRIMEarxiv}

\usepackage[utf8]{inputenc} 
\usepackage[T1]{fontenc}    
\usepackage{hyperref}       
\usepackage{url}            
\usepackage{booktabs}       
\usepackage{amsfonts}       
\usepackage{nicefrac}       
\usepackage{microtype}      
\usepackage{lipsum}
\usepackage{graphicx}
\graphicspath{{media/}}     

\usepackage{subcaption}

\usepackage{graphicx}
\usepackage{float}
\usepackage{amsmath}
\usepackage{authblk}

\title{A Novel Approach for Defect Detection of Wind Turbine Blade Using Virtual Reality and Deep Learning}
\author[1]{Md Fazle Rabbi}
\author[2]{Solayman Hossain Emon}
\author[3]{Ehtesham Mahmud Nishat}
\author[1]{Tzu-Liang (Bill) Tseng}
\author[4]{Atira Ferdoushi}
\author[5]{Chun-Che Huang}
\author[1]{Md Fashiar Rahman}

\affil[1]{Department of Industrial, Manufacturing and Systems Engineering, The University of Texas at El Paso, Texas, 79968, USA}
\affil[2]{Data Science Program, The University of Texas at El Paso, El Paso, Texas, 79968, USA}
\affil[3]{Khulna University of Engineering and Technology (KUET), Khulna-9203, Bangladesh}
\affil[4]{University of Rajshahi, Rajshahi-6205, Bangladesh}
\affil[5]{National Chi Nan University, Pu-Li, Taiwan}

\date{} 

\begin{document}
\maketitle

\thispagestyle{fancy} 

\begin{abstract}
Wind turbines are subjected to continuous rotational stresses and unusual external forces such as storms, lightning, strikes by flying objects, etc., which may cause defects in turbine blades. Hence, it requires a periodical inspection to ensure proper functionality and avoid catastrophic failure. The task of inspection is challenging due to the remote location and inconvenient reachability by human inspection. Researchers used images with cropped defects from the wind turbine in the literature. They neglected possible background biases, which may hinder real-time and autonomous defect detection using aerial vehicles such as drones or others. To overcome such challenges, in this paper, we experiment with defect detection accuracy by having the defects with the background using a two-step deep-learning methodology. In the first step, we develop virtual models of wind turbines to synthesize the near-reality images for four types of common defects - cracks, leading edge erosion, bending, and light striking damage. The Unity® perception package is used to generate wind turbine blade defects images with variations in background, randomness, camera angle, and light effects. In the second step, a customized U-Net architecture is trained to classify and segment the defect in turbine blades. The outcomes of U-Net architecture have been thoroughly tested and compared with 5-fold validation datasets. The proposed methodology provides reasonable defect detection accuracy, making it suitable for autonomous and remote inspection through aerial vehicles. 

\begin{flushleft}
\end{flushleft}

\end{abstract}

\section{Introduction}
Wind Turbine plays a vital role in the energy generation of the United States of America. Wind turbines provided almost 8.4 \% of total generated electricity in 2020 in the United States (According to US Energy Information Administration). According to the U.S Wind Turbine Database (2022), more than 700,800 wind turbines are running across 44 states resulting in the fourth number of sources for electric energy generation. Most importantly, it is the number one renewable energy source for the USA in power generation. For that reason, if the US wants to be a country of Green Energy, Wind Turbine will be their priority. As the European countries are now more focused on renewable energy for global warming and the US is lagging in this sector, the US government is now taking initiatives to grow their renewable energy source like wind turbines. Even recently, some states have set the target to maintain a percentage of energy that should come from renewable energy. So, to make the US a sustainable and low carbon emission country, wind turbine can contribute a lot. 

\newpage
\pagestyle{plain} 

Wind Turbines are generally installed in remote and rural areas where there is a constant and strong wind. The constant and strong wind can be found in both onshore and offshore locations. Hills, open fields, or coastal areas can be a very good source of wind power for onshore settlements. On the other hand, oceans or lakes are the prime sources of wind for onshore settlement. Because of wind speed and the visual impact of nature, the onshore wind turbine is preferable for installation. But both locations are remote for proper inspections of the wind turbine. After installing the wind turbine, the blade defects generally start over time. Though blade defects can occur anytime, environmental factors like lightning, flying objects, and high wind plays an important part in it. Along with that, poor inspections and maintenance are part of the gradual process of blade defects. For that reason, the detection of wind turbine defects in the initial phase is necessary for uninterrupted power generation. With the help of images of the wind turbine blades, the inspections can be done automatically and quickly by deploying deep learning-based image segmentation and classification techniques \cite{rahman2022deep, rahman2022improving}.  

 Hence, in this work, the U-Net based semantic segmentation technique is used to detect the defects in turbine blades \cite{yu2022improved}. The U-Net architecture is selected because it can give an accurate and detailed segmentation map from a small image dataset. A multilevel convolutional recurrent neural network (MCRNN) is also good for detecting wind turbine blade icing \cite{tian2021multilevel}. MCRNN is a good choice because it combines CNNs and RNNs in a hierarchical manner. In real-time supervisory control \cite{rezamand2019new} and Non-destructive testing \cite{marquez2020review}, An Unmanned Aerial Vehicle (UAV) \cite{barker2021semi} is always preferable. An UAV takes images for segmentations with the help of the embedded camera. Then those images become a great source of datasets. On the other hand, the blade defect detection's accuracy can be increased by analyzing and identifying the distinctive spectral signature \cite{rizk2021wind}. Again, there may not have enough real-world data available to train a deep learning model. Moreover, getting real-world data, in some cases, are expensive and time costuming. Therefore, it comes to the essence of using synthetic data \cite{rahman2021automatic}. As an example of using synthetic data, capturing the relationship among synthetic features \cite{abay2019privacy} and pedestrian data can be used for Deep Convolutional Neural Networks (DCNN) \cite{ekbatani2017synthetic}. And in defect detection, Fuzzy Failure Mode and Effect Analysis (FMEA) can also give a structured method for defect analysis \cite{rabbi2018assessment} or any Android application \cite{kamal2019designing}.  Infrared thermography (IRT) \cite{galleguillos2015thermographic} and continuous line laser thermography for damage visualizations for wind turbines \cite{hwang2017continuous} are other methods. However, after extensively evaluating various deep learning architectures, we have chosen U-Net as the most suitable one for this paper.  
 

\section{Problem Description}
Wind turbines show different types of defects in its blade during their lifetime. But for generalizations and understanding, in this paper, four main types of defects have been studied such as 1) Blade Crack, 2) Leading Edge Erosion, 3) Delamination, and 4) Lighting Strike Damage as shown in Figure \ref{fig:fig2}. Blade cracks are a serious issue for wind turbines. If the crack is not identified in the initial phases of its origin, the blade will eventually fail to operate. So, the identification of Blade Crack in the primary phase is a must. In this case, the proper inspection of the blade by drone or watch tower is a measure towards safe operations of wind turbines. On the contrary, leading edge Erosion is another most common defect in wind turbines as the flowing air and dust continuously cause erosions in the edges. Over time, leading edge erosion reduces the efficiency of power generation. Similarly, the delamination of wind turbines is caused by extreme weather, humidity, or mechanical stresses. This is a significant concern for the power company as it has long downtime and safety issues. The manufacturing process and materials selections play a key factor in overcoming delamination defects. Ultrasonic testing and using adhesives are part of maintenance for delamination. The lighting strike damage in wind turbines results due to the lightning effect. It may damage the generator, control systems, power electronics, or blades. The blade may be affected by direct impact or through shock waves.  

This paper addresses the problem of automated detection of the above-mentioned defects in wind turbine blades, which is a critical issue in the renewable energy industry. Traditionally, detecting blade defects is time-consuming and expensive and may not be effective in identifying all types of defects. To address this problem, the paper proposes a novel approach that combines virtual reality and deep learning techniques to develop an automated and efficient defect detection system for wind turbine blades. The proposed system aims to reduce inspection time and cost while improving the accuracy and effectiveness of defect detection from synthetic images. 

\section{Methodology}
\subsection{Designing the Wind Turbine}
To generate synthetic images, it is necessary to design a virtual model of wind turbines. The design of wind turbines can be done with the help of different Computer Aided Design (CAD) software like SolidWorks, Maya, or Blender. These are very powerful CAD software for designing any model. But SolidWorks 2021 has been chosen for its assembly tools which allow the assembly of the complex part of the model. We used SOLIDWORKS 2021 for the modeling of the wind turbine in IPS (Inchi, Pound, Second) measurements. The base of the wind turbine is made using only boss extrude tool. It has been divided the structural component mast into two sections for the convenience of the modeling both of them were made using revolve tool. Then the motor housing and the rotor are made using bose extrude tool and shell tool. It has been kept the distance between the base and the center of the rotor Z=380 in. After that, it has been used surface modeling to model one blade, and then I circular patterned the fane into 2 more blades which are equally spread into 360-degree space, centering the rotor. The Diameter of the circle made by the blade centering the rotor has become 200 in. It has been used flex tools to bend and twist the fane into 10 different possible ways to simulate the delamination possibilities of the fanes. Extrude cut has been used for lighting strike dame and blade cracks simulations in 20 different ways. To simulate the leading edge erosion, surfacing tools are used to remove some portions of the fans in 10 different ways. In total, 40 defect models have been made in this way. Finally, Keyshot 10 has been used for adding some dirt and rust material and some corrosion textures. Using Keyshot 10 all models have been exported into FBX format for further simulation in Unity which is a popular cross-platform game engine used to make virtual environment.

\begin{figure}
    \centering
    \includegraphics[scale=1.0]{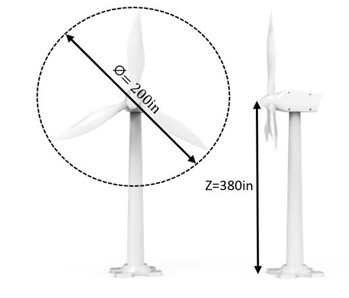}
    \caption{Designing the wind turbine}
\end{figure}

\begin{figure}[ht]
    \centering
    \includegraphics[scale=1.0]{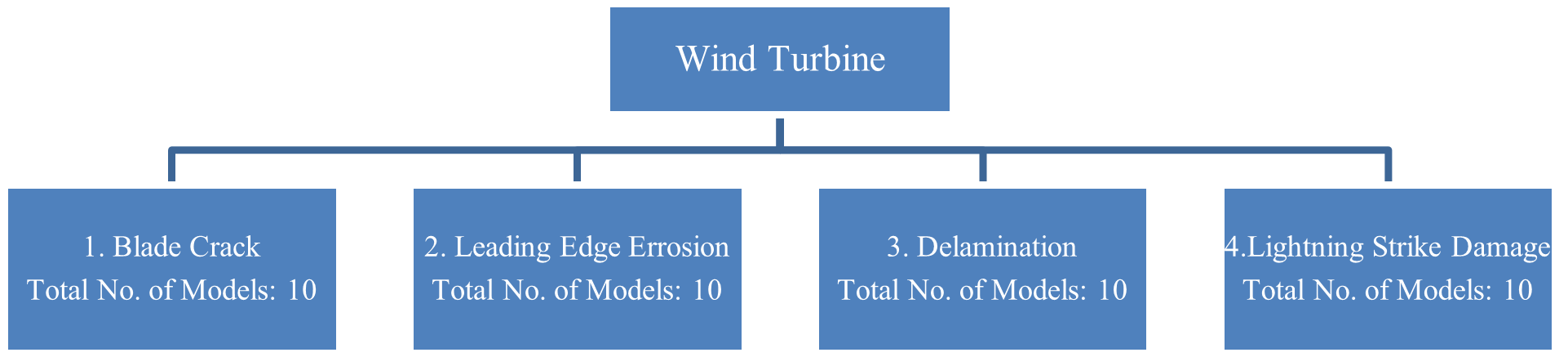}
    \caption{Schematic overview of wind turbine defect type and generated model amount}
    \label{fig:fig2}
\end{figure}

\subsection{Creating the Virtual Environment in Unity}
In this paper, the virtual world has been made inside the Unity in two different scenes: a. Terrain, b. Cube Map. In unity terrain, there are plenty of options to set up any rural environment including rocks, grass, road, and trees. Developers manipulate the vertexes of the terrain to make hills, valleys, or lowlands. In the heightmap option, a surface with a different texture can be added to the systems. For any game or simulation, the terrain is an easy way to generate the desired environment. Our terrain system has been textured with painted green grasses. Three types of ground texture have been added to the terrain to give it a realistic outlook. For the tree texture, the tree asset has been downloaded from the Unity asset store and set in the Unity terrain. The muddy road has been created with the help of different types of brushes. At last, the defective wind turbine has been imported into the scene to complete the terrain to capture images.

The second virtual scene was made by Unity Cube Map. It is used in 3D graphics or 360-degree environments. The mapped 3D geometry can complete the 3D environment. Unity cube map works on combining the top, bottom, front, back, left, and right images. Anyone can import the free version of the cube map from the online Unity Asset. The cube map was then taken as material in the scene. The defective model then has been set in the model to view and capture images from it. 

\subsection{Generating Synthetic Images using Unity Perception Package}
\begin{figure}[ht]
  \begin{minipage}[b]{0.45\textwidth}
    \centering
    \includegraphics[width=\textwidth]{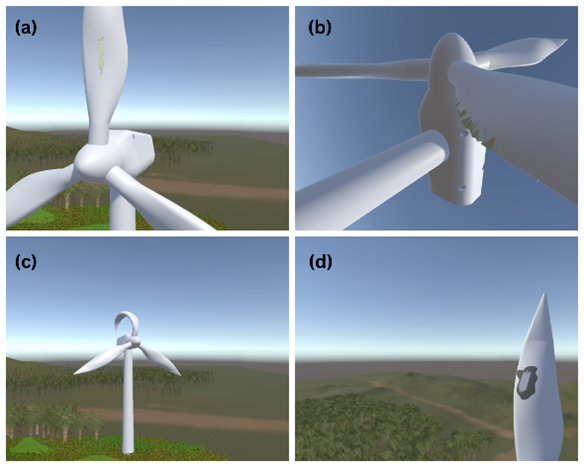}
    \caption{Generated images ((a) Blade crack, (b) Leading edge erosion, (c) Delamination, (d) Lightning strike damage)}
    \label{fig:fig3}
  \end{minipage}
  \hfill
  \begin{minipage}[b]{0.45\textwidth}
    \centering
    \includegraphics[width=\textwidth]{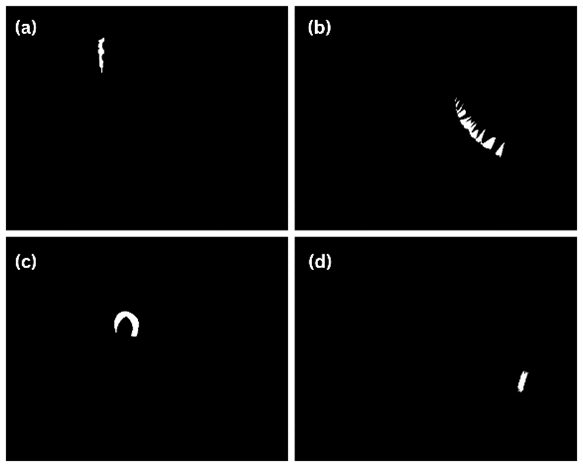}
    \caption{Generated semantic segmentation/ground truth of images ((a) Blade crack, (b) Leading edge erosion, (c) Delamination, (d) Lightning strike damage)}
    \label{fig:figure2}
  \end{minipage}
\end{figure}

The Unity Perception Package is a useful tool for capturing images from a Unity environment. To use it, you can download the package from the Unity Asset store and add it to your Unity scene. In this particular case, a defective model was placed in the scene and the camera was focused on the defective parts. Labels were added to the defective area and a Tutorial ID Label config and Semantic Segmentation Config were also included. The Semantic Segmentation Config gives the color of the defect region. The game view of Unity shows a preview of the desired capture image. This process was repeated for all 40 models, resulting in a total of 642 images and four sample can be seen Figure \ref{fig:fig3}. Additionally, all images have corresponding semantic segmentation/ground truth in the desired folders. Overall, this approach provides a way to capture images from a Unity scene and generate corresponding semantic segmentation/ground truth labels for further analysis.

\subsection{Setting up Basic U-Net Network}
\begin{figure}[H]
    \centering
    \includegraphics[scale=1.0]{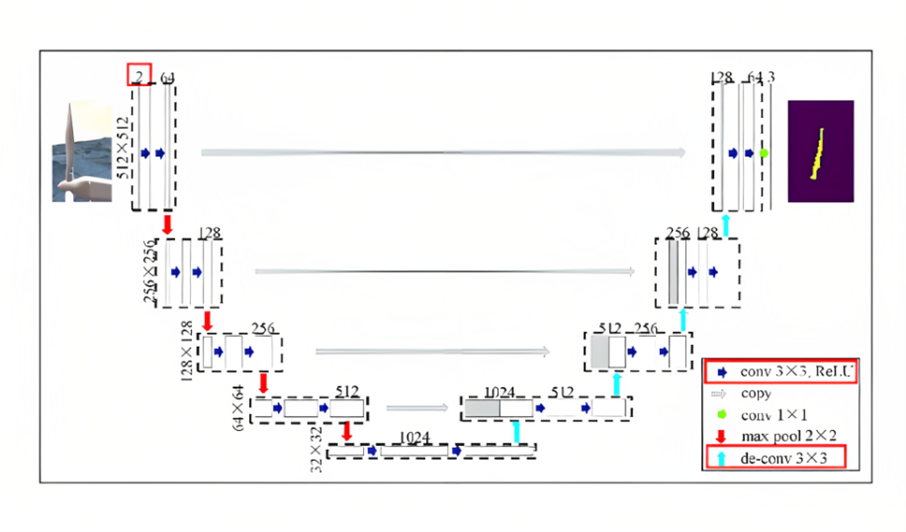}
    \caption{Illustration of U-Net convolutional network structure}
     \label{fig:fig5}
\end{figure}

We used the basic U-Net architecture \cite{fan2019ship} for image segmentation. The network consists of encoding and decoding parts that are connected by the bottleneck. The convolutional layers create the contracting paths. One or more convolutional layers make up the bottleneck, which joins the contracting and expanding paths. Its goal is to portray the most significant aspects of the input image in a compressed manner. Here, the input images are 512*512 Pixels in size. Figure \ref{fig:fig5} shows the architecture of the U-Net which is employed for our purposes. In the experimental settings, Google Colaboratory is utilized for model training and testing.

To evaluate the model outcomes, Jaccard Index is used to measure the similarity between predicted masks and ground truth masks. It measures the overlap between the ground truth and segmentation outcomes.

\begin{equation}
J(W_1, W_2) = \frac{|W_1 \cap W_2|}{|W_1 \cup W_2|}
\end{equation}

Where, $W_1$ and $W_2$ are the ground truth and segmentation outcome, respectively. The Jaccard Index ranges from 0 to 1, with 0 denoting no overlap and 1 indicating complete overlap between the sets. In the basic cross entropy loss, per-pixel loss is calculated discretely, measuring as the average of per-pixel loss. It takes into account loss in a micro sense as opposed to taking it into account globally. For handling this, we have used Dice Loss (DL) which originates from the Dice Coefficient (DSC). Dice Loss is calculated from the coefficient value by applying 1 - Dice Coefficient (DSC).

\begin{equation}
\text{DSC} = \frac{2 \times |A \cap B|}{|A| + |B|}
\end{equation}

\begin{equation}
\text{DL} = 1 - \text{DSC}
\end{equation}
Where, $A$ represents the predicted mask set and $B$ represents the ground truth mask set.

\section{Results and Discussions}
Unity generates images automatically based on the setup as discussed in Section 3. During the simulation, unity can generate tons of images within a very short period with variations with precise annotation of semantic segmentation. We use the simulated images and their corresponding annotation (ground truth) to train the U-Net model. The Adam optimizer \cite{kingma2014adam} is used to train the network. Adam is a well-known optimization algorithm for deep neural network training. Deep neural networks can be trained using this adaptive learning rate optimization approach. As the Dice Loss function gives the dissimilarity between the predicted segmentation map and the ground truth, the Dice Loss function is used here. Most importantly, here segmentations are binary, so the dice loss function selection is perfect. Finally, the Jaccard index is used as loss function. However, the total number of datasets used here are 642 with four main defect types: blade crack, leading edge erosion, delamination, and lightning strike damage. And each defect has 10 different types of variations positioning the defect in a different area and in size.

\begin{figure}[H]
  \centering
  \includegraphics[width=0.9\textwidth]{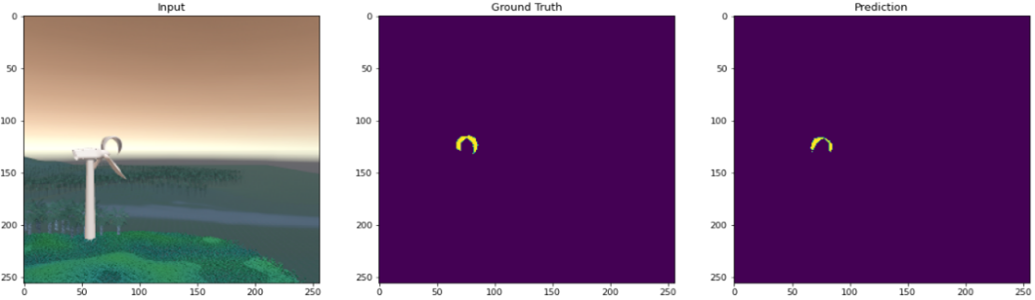}
  \label{fig:figure1}

  \vspace{\baselineskip} 

  \includegraphics[width=0.9\textwidth]{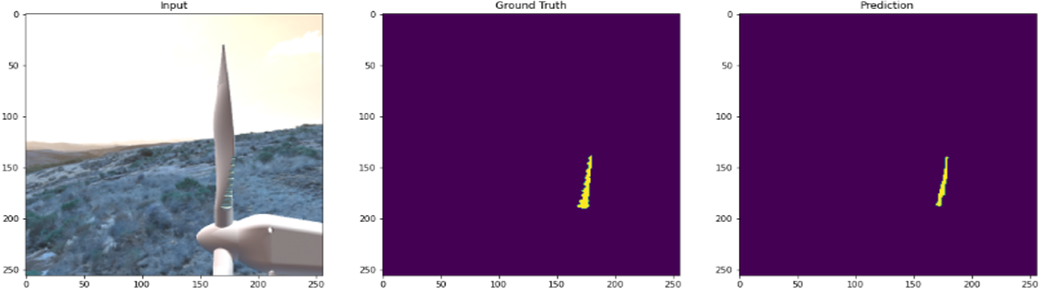}
  \label{fig:figure2}

  \caption{Visualization of Model’s Output}
  \label{fig:out}
\end{figure}

The dataset was divided into the train set (75\%) and test set (25\%) and the model was trained for 100 epochs. The improvement rate in each epoch is moderate while training the model. After training and testing the basic U-Net model, we obtain the Dice Coefficient of 0.658. A Dice Coefficient of 0.658 suggests that the U-Net model has achieved moderate segmentation accuracy, but there is still scope for improvement. In other words, the model is correctly segmenting 65.8\% of the pixels in the image. As the classes are more balanced in our experiment, the Dice Loss indicates that the model needs further refinement. Figure \ref{fig:out} also supports the finding of our experiment. As we can see in Figure \ref{fig:out} (both delamination and blade crack defects), the predicted masks from U-Net are close to the Unity-generated ground truth. But notice that the segmented result is still missing some of the pixels compared to the ground truth. Despite the limitation, the proposed methodology can segment the defects with reasonable accuracy. Table \ref{table:tab1} provides a summary of trained model and evaluation metrics.

\begin{table}[ht]
\centering
\caption{Summary of trained model and evaluation metrics.}
\begin{tabular}{|l|l|}
\hline
Dataset & 642 (images \& masks) \\
\hline
Defects & 4 types \\
\hline
Segmentation Architecture & U-Net \\
\hline
Optimizer & Adam \\
\hline
Loss Function & Jaccard Loss \\
\hline
Dice Coefficient (Test) & 0.658 \\
\hline
\end{tabular}
\label{table:tab1}

\end{table}

\section{Conclusion}
This paper focuses on data synthesis using Unity and its perception package the proposed work can successfully generate tons of images in the virtual settings. The real-like images have been used for the training of the U-Net. It is known to all that training the deep learning model needs voluminous images, where data synthesis techniques can be useful. The synthetic data were used to train the U-Net architecture. After training of 100 epochs, we obtained the Dice coefficient of 0.658. The model shows reasonable performance in detecting the detects in wind turbine blades. However, we observe some miss detection indicating the future scope to improve the outcomes. In future, we will improve the model for better segmentation and add the classification module. In addition, we will investigate the work using wide range of defects and real wind turbine images.

\bibliographystyle{unsrt}  
\bibliography{references}

\end{document}